\documentclass[a4paper, 10pt, conference]{ieeeconf}      % Use this line for a4
\usepackage{FG2019}

\FGfinalcopy % *** Uncomment this line for the final submission

\IEEEoverridecommandlockouts                              % This command is only
\usepackage{comment}

\usepackage{graphics} % for pdf, bitmapped graphics files
\usepackage{epsfig} % for postscript graphics files
\usepackage{mathptmx} % assumes new font selection scheme installed
\usepackage{times} % assumes new font selection scheme installed
\usepackage{amsmath} % assumes amsmath package installed
\usepackage{amssymb}  % assumes amsmath package installed

\usepackage{graphicx}

\usepackage{xcolor}
\usepackage{gensymb}
\usepackage{subfigure}
\usepackage{color}
\usepackage{array}
\usepackage{url}
\def\FGPaperID{93} % *** Enter the FG 2019 Paper ID here

\title{\LARGE \bf
Improving Head Pose Estimation with a Combined Loss\\ and Bounding Box Margin Adjustment
}

\author{\parbox{16cm}{\centering
    {\large Mingzhen Shao$^1$, Zhun Sun$^2$, Mete Ozay$^1$, Takayuki Okatani$^1$$^2$}\\
    {\normalsize
    $^1$ Graduate School of Information Sciences, Tohoku University, Sendai, Japan\\
    $^2$ Riken Center for AIP, Tokyo, Japan}}
}

\begin{document}
\IEEEoverridecommandlockouts\pubid{\makebox[\columnwidth]{978-1-7281-0089-0/19/\$31.00~\copyright{}2019 IEEE \hfill}
\hspace{\columnsep}\makebox[\columnwidth]{ }}

\ifFGfinal
\thispagestyle{empty}
\pagestyle{empty}
\else
\author{Anonymous FG 2019 submission\\ Paper ID \FGPaperID \\}
\pagestyle{plain}
\fi
\maketitle

%%%%%%%%%%%%%%%%%%%%%%%%%%%%%%%%%%%%%%%%%%%%%%%%%%%%%%%%%%%%%%%%%%%%%%%%%%%%%%%%
\begin{abstract}
    We address a problem of estimating pose of a person's head from its RGB image. The employment of CNNs for the problem has contributed to significant improvement in accuracy in recent works. However, we show that the following two methods, despite their simplicity, can attain further improvement: (i) proper adjustment of the margin of bounding box of a detected face, and (ii) choice of loss functions. 
    We show that the integration of these two methods achieve the new state-of-the-art on standard benchmark datasets for in-the-wild head pose estimation. The Tensorflow implementation of our work is available at \url{https://github.com/MingzhenShao/HeadPose}
    % Face pose estimation is a fundamental yet challenging task in computer vision. In the past decades, various methods have been proposed and the accuracy has got a great increase. But none of them pay attention to the background information of inputs. 
    % In this paper, we propose a parameter $K$ to control the background size and analyze the effect of the background on head pose estimation.
    % We also propose a multi-loss convolutional neural network with a new combined loss for head pose estimation. We present tests on common in-the-wild pose benchmark datasets which show state-of-the-art results.
\end{abstract}

% \addtolength{\textheight}{-3cm}   % This command serves to balance the column lengths

%%%%%%%%%%%%%%%%%%%%%%%%%%%%%%%%%%%%%%%%%%%%%%%%%%%%%%%%%%%%%%%%%%%%%%%%%%%%%%%%
\section{INTRODUCTION}
This paper considers a head pose estimation problem, which is to infer orientation of a person's head relative to the camera view from its single RGB image. Although early works can provide only a rough estimate of the head pose \cite{4497208, NG2002359,understandingpose,SHERRAH2001807, 711102, 12307}, recent methods that use an RGB-D image~\cite{12307, fanelli_IJCV, VenturelliBVC17a, 8451363} or an RGB image~\cite{6248014, Jourabloo_2016_CVPR,RanjanSCC16, RanjanPC16, KumarAC17} can estimate head pose with degree-level accuracy.
%The methods using RGB-D image require depth information which is typically obtained by a depth camera. Recently some works with this method 
The former methods 
%methods that use an RGB-D image 
are generally superior to 
the latter
%those that use an RGB image 
in terms of accuracy.
%~\cite{12307, fanelli_IJCV, VenturelliBVC17a, 8451363}. 
However, RGB-D cameras are more costly than RGB cameras and usually require active sensing, which can be inaccurate in outdoor scenes, hence limiting their applicability.
% of this method. 
%Furthermore, for applications that require immediate reactions and massive data processing within a short period of time, the use of depth camera and the additional processing time associated with it can be prohibitively expensive. 
Therefore, methods that only use an RGB image 
%serves as a compromise between speed and accuracy and 
have the potential for wider and more diverse applications. 
%Thus we consider them in this study.

There are two types of approaches to the problem of head pose estimation from an RGB image. One is to first detect multiple landmarks (key-points) of a face from the input image, then establish correspondences between them and the landmarks of a 3D generic head model, and finally estimate the relative pose of the face using the perspective-n-point (PnP) algorithm based on the established correspondence \cite{6248014, Jourabloo_2016_CVPR}. This approach is constrained by the (in)accuracy of the landmark detection, and the difference between input faces and generic head models.
The other approach is to use CNNs to directly predict the relative head pose from an input image, which is more attractive as there is no such explicit constraint on estimation accuracy. 

Several studies have been conducted on this approach so far, which have gained considerable amount of improvement in accuracy ~\cite{RanjanSCC16, RanjanPC16, KumarAC17}. However, we point out that there is still room for improvement in the current state-of-the-art. To be specific, there are two ingredients that bring about improvements. One is using margins around a bounding box of a detected face, and the other is choice of loss functions. 

\begin{figure}[tb]
\centering
\begin{minipage}[c]{0.3\linewidth}
\centering
\includegraphics[width=\linewidth]{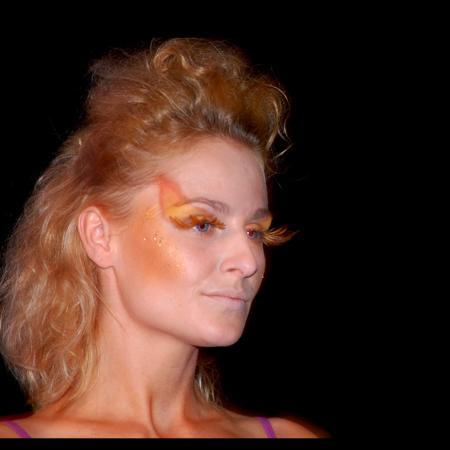}
\scriptsize Roll:18.71\\Pitch:-11.02\\Yaw:-48.37
\end{minipage}
\begin{minipage}[c]{0.3\linewidth}
\centering
\includegraphics[width=\linewidth]{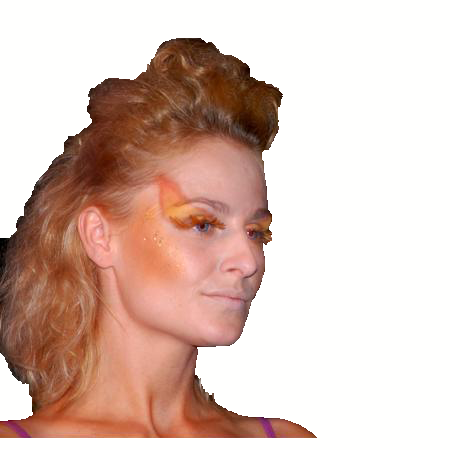}
\scriptsize Roll:22.80 \\Pitch:-7.75 \\Yaw:-48.73
\end{minipage}
\begin{minipage}[c]{0.3\linewidth}
\centering
\includegraphics[width=\linewidth]{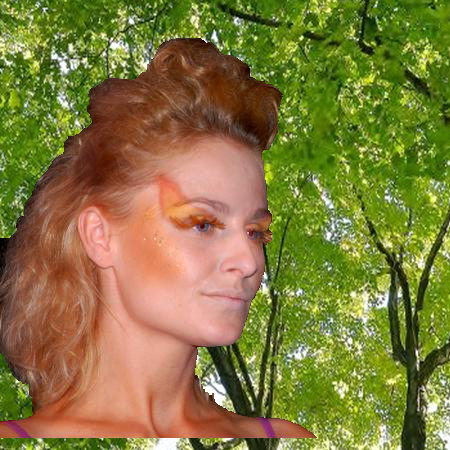}
\scriptsize Roll:23.83 \\Pitch:-9.03 \\Yaw:-50.73
\end{minipage}
\\
\begin{minipage}[c]{0.3\linewidth}
\centering
\includegraphics[width=\linewidth]{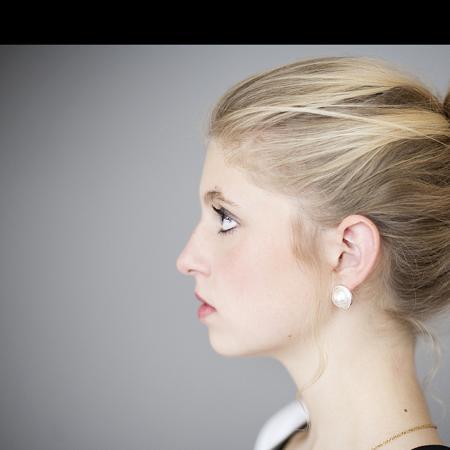}
\scriptsize Roll:3.56 \\Pitch:2.57 \\Yaw:70.43
\end{minipage}
\begin{minipage}[c]{0.3\linewidth}
\centering
\includegraphics[width=\linewidth]{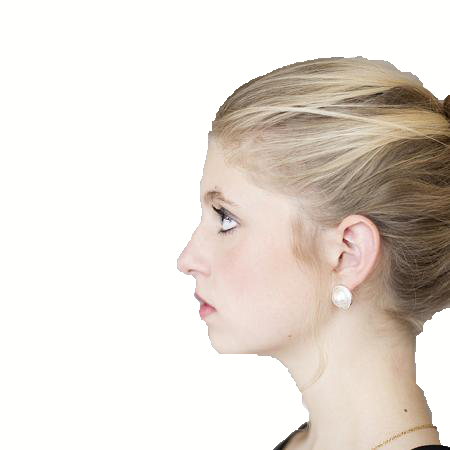}
\scriptsize Roll:10.07 \\Pitch:6.13 \\Yaw:78.26
\end{minipage}
\begin{minipage}[c]{0.3\linewidth}
\centering
\includegraphics[width=\linewidth]{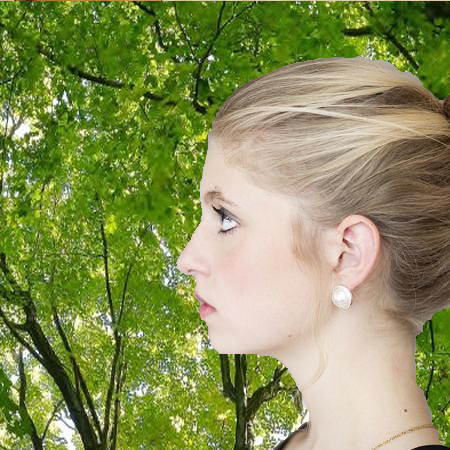}
\scriptsize Roll:6.48 \\Pitch:2.34 \\Yaw:73.17
\end{minipage}
\newline
\caption{Examples showing the effect of different types of background on pose estimation.}
\label{I1}
\end{figure}

We have discovered in our experiments that bounding box margin has a large impact on the final accuracy of head pose estimation. In other words, head pose estimation is vulnerable to changes in the background scene around the target face, as shown in Fig.\ref{I1}, although this may not be so surprising; the methods are appearance-based after all. However, we show that proper adjustment of the bounding box margin mitigates this vulnerability. Furthermore, we have explored the space of loss functions for face pose estimation, having found a better choice that contributes to further improvement. The reason why these ingredients are overlooked in previous studies may be because the head pose estimation task was treated as an auxiliary task (or by-product) of other main tasks, such as face alignment, face recognition and detection; this may make it hard to think about bounding box margins and different loss functions. 

% {\color{green}
% In our training process, we find that the content of the background and the size of the background all influence the final prediction result, as shown in Figs~\ref{I1} and \ref{I2}. So we propose a parameter $K$ to control the background size in order to get a better result.
% }

The contributions of this paper are summarized as follows.
\begin{itemize}
    \item It is empirically analyzed how the size of the bounding box margin affects the final pose estimation. 
    %\item We propose a new one-stage method for head pose Euler angles estimation directly from image intensities using a combined loss network.
    \item A new combined loss is proposed, and we confirmed in the experiments that our proposed loss is superior to the normal MSE loss which is widely employed in previous studies. 
    \item The combination of the above two has achieved the state-of-the-art performance on the public benchmark dataset AFLW2000 and BIWI.
\end{itemize}

%%%%%%%%%%%%%%%%%%%%%%%%%%%%%%%%%%%%%%%%%%%%%%%%%%%%%%%%%%%%

\begin{figure}[tb]
    \centering
    \includegraphics[width=0.8\linewidth]{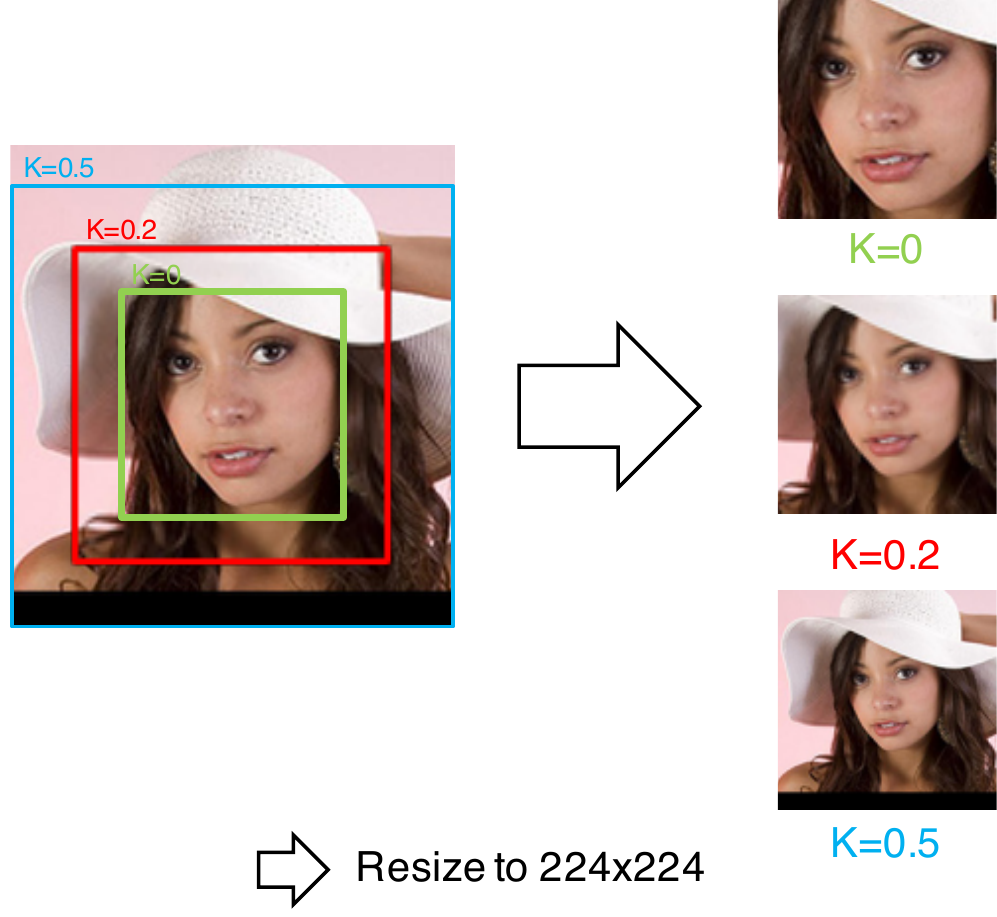}
    \caption{Adjustment of the margin of a bounding box. A face detector provides a bounding box, whose margin is controlled as shown above by a control parameter $K$; the box with the margin is cropped and fed into our CNN.}
    \label{fig:my_label}
\end{figure}

\section{PROPOSED METHOD}

% {\color{green}
% In this section, we first describe our proposed CNN. Next, we introduce how   angles of head pose are represented in our work. Then, we introduce our proposed combined loss function utilized for training CNNs to learn  representations of pose. At last, we suggest using a cropping coefficient $K$ that controls the size of area of images fed into the proposed CNN.
% The flowchart of our method is shown in Fig.~\ref{net}.
% }

Our proposal consists of two methods as stated above, i.e.,  adjustment of the margin of a bounding box of a face that is inputted to our CNN and a new loss for training it along with a novel output layer. We first explain the latter and then the former in what follows. 

\begin{figure}[tb]
    \centering
    \resizebox{.9\linewidth}{!}{
    \includegraphics{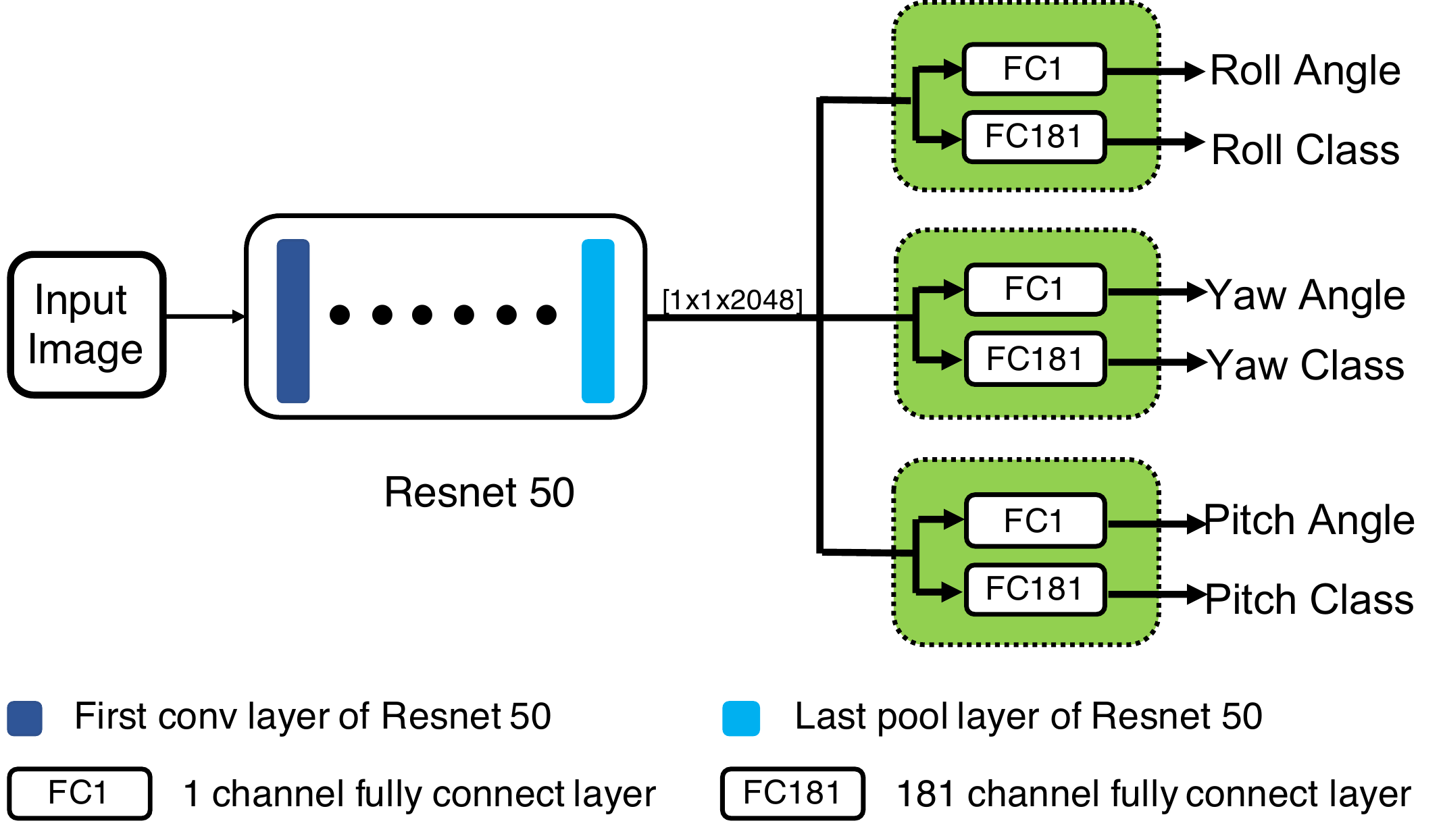}
    }
    \caption{The architecture of the proposed CNN.}
    \label{structure}
\end{figure}

\subsection{Losses and Network Design}

Our CNN receives a three-channel (RGB) image of a person's head as an input, and outputs the pitch, roll, and yaw angles of the head; see Fig.~\ref{structure}.
Employing a ResNet50~\cite{He2015} pretrained on the ImageNet dataset for the backbone of our CNN, 
we place three blocks of layers on top of its last average pooling layer. Each block outputs a prediction of one of the roll, yaw, and pitch angles.

In previous studies, the problem of head pose estimation is usually posed as regression, in which each rotation angle is predicted. The sum of squared difference between the true and predicted rotation angles is used for a regression loss to be minimized in training. In this study, we propose to use a classification loss in addition to the regression loss. To be specific, we split a set of degrees of angles which take value in the interval [-90\degree:90\degree] into 181 classes uniformly; each class corresponds to one-degree range. Then, our CNN predicts an angle with its discrete class and a continuous number for each of roll, yaw, and pitch angles, as shown in Fig.~\ref{structure}. Each of the three head blocks receives the same output (2048 floating point numbers) from the average pooling layer of the ResNet50, and outputs a predicted angle in a continuous number as well as the scores of the 181 angle classes. For the former, we use a {$2048\times 1$} fully-connected layer that maps the output obtained from the backbone ResNet50 to a single continuous number. For the latter, we use a {$2048\times 181$} fully-connected layer that maps the same backbone output followed by a softmax function to obtain 181 class scores. 

To train the CNN, we use a combined loss function for each angle in the following way. The regression loss ${L}_{MSE}$ is the mean squared error computed over the training samples ($i=1,2,\ldots,n$) by
\begin{equation}
    {L}_{MSE} = \frac{1}{n}\sum_{i=1}^{n}(y_i - \hat{y_i})^2,
\end{equation}
where $y_i$ is the true angle and $\hat{y}_i$ is the predicted angle for the $i$-th sample. 
For the classification loss ${L}_{S}$, we employ the temperature scaling to make class scores distribute wider by
\begin{equation}
    {L}_{S} = -\frac{1}{n}\sum_{i=1}^{n}\log
    \frac{\exp{\left(
    %\frac{W_{y_i}^\top f_i}{T}
    W_{y_i}^\top \hat{y}_i / T
    \right)}}
    {\sum^{181}_{j=1}\exp{\left(
    %\frac{W^\top_jf_i}{T}
    W^\top_j\hat{y}_i/T
    \right)}},
\end{equation}
where $W_j$ is the $j$-th column of the last fully connected layer and  $\hat{y}_i$ is the input to it for the $i$-th sample; $y_i$ is the true class for the $i$-th sample; $T$ is the temperature scaling parameter, which is set to $2$ throughout our experiments. We then add the two losses to compute the final loss function used in training by
\begin{equation}
    {L} = {L}_{S}(y,\hat{y}) + \alpha {L}_{MSE}(y,\hat{y}),
\end{equation}
where $\alpha$ is the weight balancing the two losses. We set $\alpha$ to be 2 throughout our experiments. 

The design of our CNN and the loss functions explained above is based on our conjecture that the additional employment of the classification loss will guide the CNN to attain better global optimum. 

% The final loss is,
% \begin{equation}
%     \mathcal{L}_{ALL} = \mathcal{L}_{Roll} + \mathcal{L}_{Pitch} + \mathcal{L}_{Yaw}
% \end{equation}

% \begin{equation}
%     \mathcal{L}_{AMS} = -\frac{1}{n}\sum_{i=1}^{n}\log\frac{e^{s\cdot(W_{y_i}^{T}f_i-m)}}{e^{s\cdot(W^{T}_{y_i}-m)}+\sum^c_{j=1,j\neq{y_i}}e^{sW^T_jf_i}}
% \end{equation}

\subsection{Adjustment of Bounding Box Margin}

Given an input image $X$, we apply a face detector to it to obtain a square bounding box of a face. We denote the coordinates of the detected bounding box by ${[t_x,t_y]-[t_x+t_l,t_y+t_l]}$ (top-left to bottom-right corners), where $t_{l}$ is the size of the square bounding box. 
%Assume the top-left corner of the bounding box in the image as the origin of a pixel coordinate system, whose x-axis and y-axis is defined from left-to-right and top-to-bottom as $t_x$ and $t_y$. Then we get a face detection bounding box $[t_x, t_y, t_l]$. 
Given this bounding box, we crop the image with additional margins, which will be inputted to our CNN. We denote the coordinates of the crop box with additional margins by $[t_{x} - K t_{l}, t_y - K t_{l}]-[t_x + t_l + K t_l, t_y + t_l + K t_l]$, where $K$ is a control parameter. 

% Our method can be used together with any face detector.
For the face detector, we can use any method having sufficient accuracy. In the following experiments, we used the method of Chen et al.~\cite{dong}. We also tested Microsoft Face API\footnote{https://azure.microsoft.com/en-us/services/cognitive-services/face/} and confirmed that it provided very similar results. 

\section{EXPERIMENTAL RESULTS}

We conducted several experiments to test the proposed approach.

\subsection{Datasets}

Several different datasets have been developed so far for head pose estimation. As it is difficult to precisely measure (or manually annotate) the 3D pose of a head, most of them generate the ``ground truth'' head poses by fitting a mean 3D face with the POSIT algorithm. Although this approach provides accurate results for small angle head poses, it can only provide sub-optimal results for large angle head poses, because the accuracy of landmarks detection will deteriorate for them. We confirmed this tendency in our preliminary experiments. Thus, instead of these datasets developed for head pose estimation, we choose 300W-LP~\cite{ZhuLLSL15} as a training dataset,which contains 61,255 images of faces having large poses, which are synthetically generated from the 300W~\cite{30_w} dataset in such a way that depth of each face is first predicted and then its profile views are generated with 3D rotation. 

We choose AFLW2000~\cite{ZhuLLSL15} for a test dataset, which contains 2,000 identities of the AFLW dataset that have been re-annotated with sixty-eight 3D landmarks using a 3D model. It covers head poses with large variations, different illumination and occlusion conditions, and thus it can be used as a prime target for evaluation of head pose estimators. 
We also test our method on some datasets captured by RGB-D cameras like BIWI~\cite{fanelli_IJCV} and SASE~\cite{sase}. 
The BIWI dataset which captured in a laboratory setting contains 24 videos with 15,678 frames, generated from RGB-D videos captured by a Kinect device for different subjects and head poses. A 3D model is fitted to the RGB-D videos to track and obtain the ground truth head poses. The range of the head pose angle is $\pm77\degree$ for yaw, $\pm60\degree$ for pitch and $\pm50\degree$ for roll. {SASE use the same method but applie Kinect 2 camera which offers a higher quality of depth information.}

\subsection{Training Setting}
 
We first apply the face detector to all the training and testing images to get the bounding box of each face. We crop the bounding box with margin specified by $K$ as above, and resize it to $224\times 224$, which fits to the input size of our CNN. Using batch size 64, 
%and the label contains three channels each combined with a Euler angle and a classification label. 
we train our CNN for 100 epochs on the 300W-LP dataset using stochastic gradient descent (SGD) optimizer with the learning rate of $10^{-4}$ and momentum parameter of $0.9$.

\subsection{Comparison with State-of-the-Art Methods}

We first show comparison of our method with the state-of-the-art methods on AFLW2000 and BIWI datasets. For the AFLW2000 dataset, we remove 36 images that have angles larger than $90\degree$, because we only consider the head pose which takes values in $\pm90\degree$.

\begin{table}[h!]
    \centering
    \caption{Average error of Euler angles across different methods on the AFLW2000 dataset.}
    \resizebox{\linewidth}{!}{
    \begin{tabular}{ |c||c|c|c|c| }
         \hline
        % \multicolumn{4}{|c|}{Country List} \\
        % \hline
          & Yaw & Pitch & Roll & MAE\\
         \hline
         HopeNet($\alpha$=2)~\cite{00925}& $6.470$ & $6.559$ & $5.436$ & $6.155$\\
         3DDFA~\cite{ZhuLLSL15} & $5.400$ & $8.530$ & $8.250$ & $7.393$\\
         FAN~\cite{BulatT17a} & $6.358$ & $12.277$ & $8.714$ & $9.116$\\
         Dlib~\cite{Kazemi2014OneMF} & $23.153$ & $13.633$ & $10.545$ & $15.777$\\
         Two-Stage~\cite{8099876} & $11.917$ & $8.252$ & $7.471$ & $9.213$\\
         Ours~($K$ = 0.5) & $\textbf{5.073}$ & $\textbf{6.369}$ & $\textbf{4.992}$ & $\textbf{5.478}$ \\
         \hline
        \end{tabular}
        }
        \newline
    
    \label{T1}
\end{table}

Table~\ref{T1} shows results of our approach and the state-of-the-art methods on the AFLW2000 dataset. It is seen that the proposed method achieves the best accuracy for all yaw, pitch and roll angles by a large margin as compared with others. The two-stage methods (i.e., those that detect and use landmarks) tend to yield large errors due to failure of landmark detection. The direct methods including ours predict angles directly from image intensities and tend to provide better performance. Table~\ref{T2} shows results on the BIWI dataset. Although it yields slightly less accurate estimate for the roll angle estimation, the proposed method outperforms others by a large margin in the mean absolute error (MAE).
\begin{table}[h!]
    \centering
    \caption{Average error of Euler angles across different methods on the BIWI dataset.
    % ${^\ast}$Theses methods use depth information.
    }
    \resizebox{\linewidth}{!}{
        \begin{tabular}{ |c||c|c|c|c| }
         \hline
        % \multicolumn{4}{|c|}{Country List} \\
        % \hline
          & Yaw & Pitch & Roll & MAE\\
         \hline
        %  OpenFace\cite{Baltrusaitis2016}${^\ast}$ & $7.800$ & $8.000$ & $4.600$ & $6.800$\\
        %  3DMM+FHM\cite{7961811}${^\ast}$ & $2.500$ & $1.500$ & $2.200$ & $2.067$\\
        %  \hline
         KEPLER~\cite{KumarAC17} & $8.084$ & $17.277$ & $16.196$ & $13.852$\\
         FAN~\cite{BulatT17a} & $8.532$ & $7.483$ & $7.631$ & $7.882$\\
         Dlib~\cite{Kazemi2014OneMF} & $16.756$ & $13.802$ & $6.190$ & $12.249$\\
         3DDFA~\cite{ZhuLLSL15} & $36.175$ & $12.252$ & $8.776$ & $19.068$\\
         Two-Stage~\cite{8099876} & $9.488$ & $11.339$ & $\textbf{6.002}$ & $8.943$ \\
         Ours~($K$ = 0.5) & $\textbf{4.592}$  & $\textbf{7.254}$ & $6.150$ & $\textbf{5.999}$ \\
         \hline
        \end{tabular}
        }
        \newline
    \label{T2}
\end{table}

% Since the BIWI dataset contains depth information, we also include two methods based on RGB-D image, OpenFace~\cite{Baltrusaitis2016} and 3DMM+FHM~\cite{7961811}. Our result outperforms OpenFace and is closing the gap between 3DMM-FHM, which is one of the state-of-the-art methods that uses depth information.

\subsection{Effects of Bounding Box Margins}

We show that the choice of the margin parameter $K$ has a significant impact on accuracy in the proposed method. We evaluated the performance of our method by varying $K$ in the interval from $0.0$ (the original bounding box) to $1.0$. To be specific, 
we train the proposed CNN on the 300W-LP dataset for 100 epochs and test it on the AFLW2000 dataset. As shown in Table~\ref{T3}, 
the accuracy attains the best around the intermediate values between $0.0$ and $1.0$. Based on this observation, we can state that $K = 0.5$ provides the best result. 
\begin{table}[h]
    \centering
    \caption{Average error of Euler Angles obtained using the combined loss with different $K$.}
    \resizebox{\linewidth}{!}{
    \begin{tabular}{|c||c|c|c|c|}
         \hline
         & Yaw & Pitch & Roll & MAE\\
         \hline
         Combined loss ($K = 0.0$)& $5.773$ & $6.720$ & $5.357$ & $5.950$\\
         Combined loss ($K = 0.2$)& $5.082$ & $6.470$ & $4.850$ & $5.467$\\
         Combined loss ($K = 0.3$)& $5.097$ & $6.223$ & $\textbf{4.727}$ & $5.350$\\
         Combined loss ($K = 0.4$)& $4.850$ & $6.275$ & $5.019$ & $5.382$\\
         Combined loss ($K = 0.5$)& $\textbf{4.749}$ & $\textbf{6.191}$ & $4.764$ & $\textbf{5.234}$\\
         Combined loss ($K = 0.6$)& $4.976$ & $6.397$ & $4.902$ & $5.425$\\
         Combined loss ($K = 1.0$)& $4.866$ & $7.140$ & $5.075$ & $5.693$\\
         \hline
    \end{tabular}
    }
    \newline
    \label{T3}
\end{table}

\begin{figure}[tb]
    \centering   
    \includegraphics[width=\linewidth]{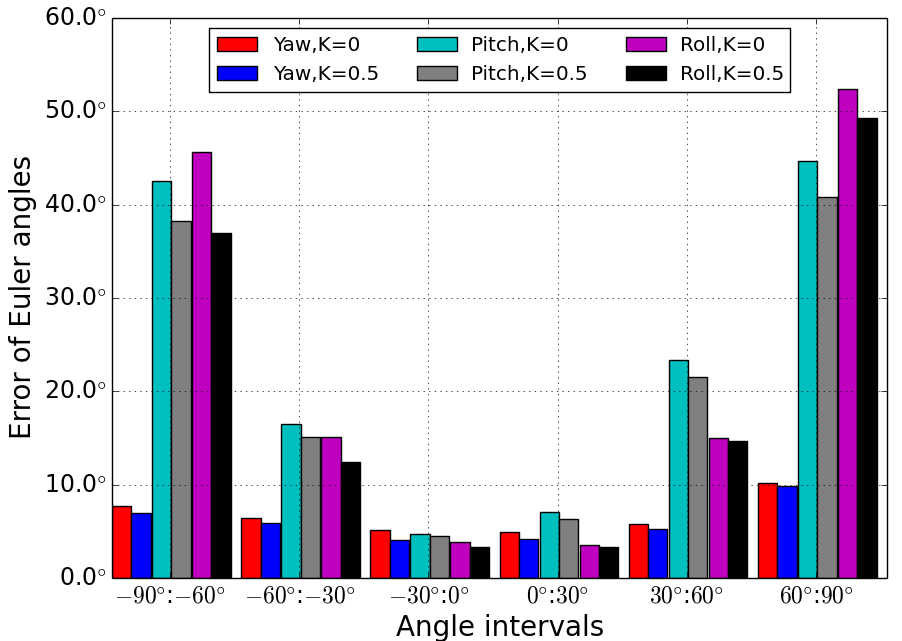}
    % \begin{minipage}[c]{\linewidth}
    % \centering
    % \includegraphics[width=\linewidth]{FG2019/imgs/Yaw.pdf}
    % \end{minipage}
    % \begin{minipage}[c]{\linewidth}
    % \centering
    % \includegraphics[width=\linewidth]{FG2019/imgs/Pitch.pdf}
    % \end{minipage}
    % \begin{minipage}[c]{\linewidth}
    % \centering
    % \includegraphics[width=\linewidth]{FG2019/imgs/Roll.pdf}
    % \end{minipage}
    % \caption{The relation between K and MAE}
    \caption{Average estimation errors obtained using two different margin parameters $K$ for different angle values on the AFLW2000.}
    \label{K_effect}
\end{figure}
Fig.~\ref{K_effect} shows detailed behaviour of the proposed method for different angle values with two $K$ values ($0.0$ and $0.5$). 
It is observed that the yaw angle is accurate for a wide range of angles, whereas the pitch and roll angles tend to be inaccurate for larger (absolute) angles. It is then observed that the use of the bounding box margin ($K=0.5$) improves accuracy for all three angles over all angle values. The improvement tends to be larger for the cases where we have large errors, i.e., large pitch and roll angles. 

% We note that the estimation error of Yaw channel is not change greatly with the head pose variation. The parameter $K$ improve the performance equally on different degree.
% On the Pitch and Roll channels, the estimation is accurate at a small range of the head pose. While the pose enlarges, the error increases hugely. With the introduction of parameter $K$, the estimation accuracy on large pose has efficient increase. 
% Which prove the background
% % provide extra information for head pose estimation and it
% has more important effect when the prediction result has a huge error with the ground-truth.

\subsection{Comparison with a Standard Regression Loss}

\begin{figure}[tb]
    \centering
    \includegraphics[width=\linewidth]{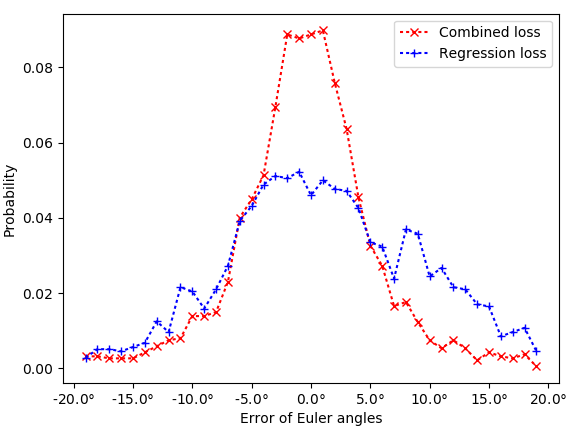}
    \caption{Distribution of estimation error on yaw angle for the combined loss and regression loss.}
    \label{constraint}
\end{figure}

We also conducted an ablation test with respect to the proposed combined loss. Specifically, we compare the performance of the proposed method with and without using the classification loss; the proposed loss combined without using the classification loss is identical to the standard regression loss ($L2$ loss) used in \cite{RanjanSCC16, RanjanPC16, KumarAC17}. Thus, we train our CNN with and without using the classification loss for several $K$ values on the 300W-LP dataset for 100 epochs, and then test it on the AFLW2000 dataset. The results given in Table~\ref{T4} show that for all $K$ values, the combined loss provides better accuracy than using just the regression loss. We conjecture that this is because of the fact that the classification loss provides a stronger constraint especially in the range of small errors, contributing more robust back-propagation of gradient compared to using the regression loss alone. 
{
% \color{blue}
Taking yaw angle as an example, we demonstrate the distribution of the estimation error for the combined loss and the standard regression loss in Fig.~\ref{constraint}. It is observed that 
more samples distribute near $0.0\degree$ by using combined loss 
% the variance of distribution using combined loss is less than the distribution using regression loss 
which shows that the network trained with combined loss is more robust than the network trained with regression loss.}
%The idea behind this combined loss is that by performing bin classification, we use the very stable softmax layer and cross-entropy, thus the network learns to predict the neighborhood of the pose in a robust fashion. 

\begin{table}[h]
    \centering
    \caption{Average estimation error obtained using the combined loss (regression and classification losses) and the regression loss alone.}
    \resizebox{\linewidth}{!}{
    \begin{tabular}{|c||c|c|c|c|}
         \hline
         & Yaw & Pitch & Roll & MAE\\
         \hline
         Combined loss ($K = 0.0$)&$5.773$ & $6.720$ & $5.357$ & $\textbf{5.950}$\\
         Regression loss ($K = 0.0$)& $6.007$ & $6.860$ & $5.540$ & $6.136$\\
         \hline
         Combined loss ($K = 0.25$)& $5.276$ & $6.504$ & $4.905$ & $\textbf{5.562}$\\
         Regression loss ($K = 0.25$)& $5.395$ & $6.653$ & $5.072$ & $5.707$\\
         \hline
         Combined loss ($K = 0.5$)& $4.749$ & $6.191$ & $4.764$ & $\textbf{5.234}$\\
         Regression loss ($K = 0.5$)& $5.121$ & $6.625$ & $5.170$ & $5.639$ \\
         \hline
        
    \end{tabular}
    }
    \newline
    \label{T4}
\end{table}

%%%%%%%%%%%%%%%%%%%%%%%%%%%%%%%%%%%%%%%%%%%%%%%%%%%%%%%%%%%%%%%%%%%%%%%%%%%%%%%%
\section{CONCLUSION}

In this paper, we have proposed two methods for improving head pose estimation with a CNN that directly predicts the yaw, pitch and roll angles from a single RGB image. We show that the bounding box margin has impact on estimation accuracy, and also show a new combined loss function of a standard $L2$ regression loss and a classification loss. We show through experiments that these two contribute to improve the state-of-the-art on several public benchmark datasets.

\section*{Acknowledgments} 

This work was partly supported by JSPS KAKENHI Grant Number JP15H05919 and JST CREST Grant Number JPMJCR14D1.

%%%%%%%%%%%%%%%%%%%%%%%%%%%%%%%%%%%%%%%%%%%%%%%%%%%%%%%%%%%%%%%%%%%%%%%%%%%%%%%%

% \begin{thebibliography}{99}

% \bibitem{c1}
% J.G.F. Francis, The QR Transformation I, {\it Comput. J.}, vol. 4, 1961, pp 265-271.

% \bibitem{c2}
% H. Kwakernaak and R. Sivan, {\it Modern Signals and Systems}, Prentice Hall, Englewood Cliffs, NJ; 1991.

% \bibitem{c3}
% D. Boley and R. Maier, "A Parallel QR Algorithm for the Non-Symmetric Eigenvalue Algorithm", {\it in Third SIAM Conference on Applied Linear Algebra}, Madison, WI, 1988, pp. A20.
% \bibliography{reference}

% \end{thebibliography}

{\small
\bibliographystyle{ieee}
\bibliography{reference}
}

\end{document}